\documentclass{article}
\usepackage{spconf}

\usepackage{graphicx}
\usepackage{amssymb}
\usepackage{amstext}
\usepackage{amsmath}
\usepackage{cite}
\usepackage{hyperref} 
\usepackage{balance}

\title{MorphSet: Augmenting categorical emotion datasets \\with dimensional affect labels using face morphing}

\twoauthors
  {Vassilios Vonikakis \thanks{Part of this work was conducted while V. Vonikakis and S. Winkler were with the Advanced Digital Sciences Center (ADSC), University of Illinois at Urbana-Champaign, Singapore. The contribution from V. Vonikakis was made by him prior to joining AWS. Contact: winkler@comp.nus.edu.sg}}
	{Amazon Web Services\\
	Singapore\\
	}
  {Neo Yuan Rong Dexter, Stefan Winkler }
	{National University of Singapore\\
    School of Computing\\
	}

\begin{document}
%
\maketitle
\begin{abstract}
Emotion recognition and understanding is a vital component in human-machine interaction. Dimensional models of affect such as those using valence and arousal have advantages over traditional categorical ones due to the complexity of emotional states in humans. However, dimensional emotion annotations are difficult and expensive to collect, therefore they are not as prevalent in the affective computing community. To address these issues, we propose a method to generate synthetic images from existing categorical emotion datasets using face morphing as well as  dimensional labels in the circumplex space with full control over the resulting sample distribution, while achieving augmentation factors of at least 20x or more.
\end{abstract}
%
%
\section{Introduction}
\label{sec:intro}



Classification of basic prototypical high-intensity facial expressions is an extensively researched topic. Inspired initially by the seminal work of Ekman \cite{ekman1}, it has made significant strides in recent years \cite{review1,DL_FEA}. However, such approaches have limited applicability in real life, where people rarely exhibit high-intensity prototypical expressions; low-key, non-prototypical expressions are much more common in everyday situations. Consequently, researchers have started to explore  alternative approaches, such as intensity of facial action units \cite{Rudovic,AUintensity}, compound expressions \cite{compound}, or dimensional models of facial affect \cite{circumplex,affectnet,Aff-Wild}. Yet these alternatives have not received much attention in the computer vision community compared to categorical models.


One major problem that impedes the widespread use of dimensional models is the limited availability of datasets. This stems from the difficulty of collecting large sets of images across many subjects and expressions. It is even more difficult to acquire reliable emotion annotations  for supervised learning. Continuous dimensional emotion labels such as Valence and Arousal are difficult for laymen users to assess and assign, and hiring experienced annotators to label a large corpus of images is prohibitively expensive and time consuming. Since even experienced annotators may disagree on these labels, multiple annotations per image are required, which further increases the cost and complexity of the task. Yet there are no guarantees that the full range of possible expressions and intensities will be covered, resulting in imbalanced datasets. Consequently, large, \emph{balanced} emotion datasets, with high-quality annotations, covering a wide range of expression variations and expression intensities of many different subjects, are in short supply. 

Our approach attempts to address this need. We propose a fast and cost-effective augmentation framework to create balanced, annotated image datasets, appropriate for training Facial Expression Analysis (FEA) systems for dimensional affect. The framework uses high-quality facial morphings to transform typical categorical datasets (usually 7 expressions per subject) into dimensional ones, with an augmentation factor of at least 20x or more.  More importantly, it produces \emph{multiple annotated expressions per subject}, balanced across the Valence-Arousal (VA) space. The resulting synthesized facial images look realistic and visually convincing. We also demonstrate that they can be used very effectively for training and testing real-world FEA systems. 

Although morphing, as a means of deriving an extended set of facial expressions, is a widely used tool in psychology \cite{psychology1, psychology4}, it has found limited adoption in the computer vision community. Traditional work on expression synthesis usually incorporates manipulation of the facial geometry and texture mapping in images or videos \cite{expr_synth_review, deformable-ibug}.  
Other approaches include the use of 3D meshes adopted from RGB-D space \cite{rgbd1} or 
the adjustment of Action Units from the Facial Action Coding System (FACS) \cite{FACS_expression_synthesis}. 
More recently, researchers have employed Generative Adversarial Networks (GANs) for this purpose \cite{GAN_face_review}. Various conditional GAN variations have been used to generate novel expressions while preserving identity and other facial details \cite{exprGAN,UCGAN, slidergan}. While most of them also take the categorical approach, a few models have been proposed based on action units \cite{GANnimation} and continuous emotion dimensions \cite{many_moods_of_emotion}.  
These GANs generally require a large dataset to start with, with no guarantees that generated faces will not exhibit unnatural artifacts, and the difficulty of creating proper annotations remains. In our approach on the other hand, we have full control over the pipeline, resulting in deterministic outputs both in terms of synthetic images and dimensional emotion labels. 



Our main contributions can be summarized as follows:
\begin{itemize}
\item A new dataset augmentation framework that can transform a typical categorical facial expression dataset into a \emph{balanced} augmented one.
\item The framework can generate hundreds of different expressions \emph{per subject} with full user control over their distribution.
\item The augmented dataset comes with automatically generated, highly consistent Valence/Arousal annotations of continuous dimensional affect.  
\end{itemize}
The code for the proposed augmentation framework is available at
\href{https://github.com/dexterdley/MorphSet}{https://github.com/dexterdley/MorphSet}.

\section{Dataset Generation}
\label{sec:augmentation}
We assume a 2-dimensional \emph{polar} affective space, similar to the Valence-Arousal (VA) space of the circumplex model \cite{circumplex}, with Neutral located at the center. The typical 7 facial expressions, which are usually included in categorical emotion datasets, Neutral (Ne), Happy (Ha), Surprised (Su), Afraid (Af), Angry (An), Disgusted (Di) and Sad (Sa), can be mapped to points with specific coordinates in the polar AV space. Apart from these 7 points however, there is a lot of empty space on the remaining AV plane. These missing facial expressions comprise: (a) different \emph{expression variations} e.g.\ Delighted, Excited, Upset etc., located at different angles in the AV polar space, and (b) different \emph{intensity variations} of the expressions, e.g.\ slightly happy, moderately happy, extremely happy etc., spanning the area from the center (Neutral) outwards toward the periphery of the AV space. The basic premise of MorphSet is that many of these expression variations can be synthesized by high-quality morphings between images of categorical expressions.

\begin{figure*}[t]
  \centering
  \includegraphics[width=0.98\textwidth]{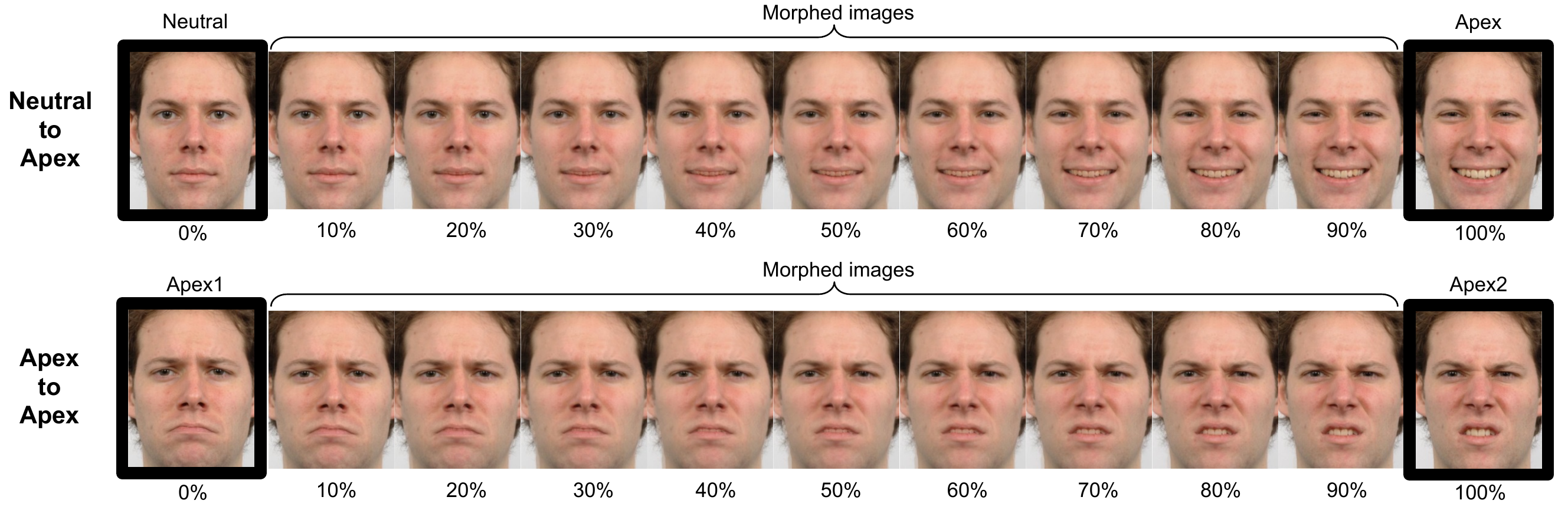}
	\caption{Examples of the 2 types of face morphings utilized in the proposed augmentation framework, using images from the Radboud dataset \cite{rafd}. In this example, all images are synthesized out of 4 given images from the original dataset (outlined in black). Top: Neutral to Apex (Happy) morphing. Bottom: Apex1 (Sad) to Apex2 (Disgusted) morphing.}
    \label{fig:morphing_examples}
\end{figure*}

Let $\mathbf{F}_{i}^{E}$ denote the face image of subject $i$ with facial expression $E$. For categorical datasets, usually $E\in \lbrace \textrm{Ne,Ha,Su,Af,An,Di,Sa} \rbrace$. Let $\theta^{E}$ denote the specific angle of each expression in the polar AV space, estimated from emotion studies  \cite{circumplex, paltoglou}. Let $I_{i}^{E} \in [0,1]$ denote the intensity of expression $E$ of subject $i$. Zero expression intensity $I^{E}=0$ coincides with Neutral (by definition $I^{\textrm{Ne}}=0$), while $I^{E}=1$ represents the highest possible intensity. 

Let $M_{p}\left(\mathbf{F}_{i}^{source}, \mathbf{F}_{i}^{target}, r\right)$ be a morphing function, based on $p$ facial landmarks, that returns a new face image, which is the result of morphing $\mathbf{F}_{i}^{source}$ towards $\mathbf{F}_{i}^{target}$ with a ratio $r\in \left[0,1\right]$; when $r=0$ the morphed image is identical to $\mathbf{F}_{i}^{source}$, and when $r=1$ it is identical to $\mathbf{F}_{i}^{target}$. Any contemporary morphing approach can be used for this, such as Delaunay triangulation followed by local warping of 68 facial landmarks from Dlib \cite{dlib09} face recognition system. 

Our augmentation framework is based on 2 types of morphings. In order to synthesize new expression variations, Apex to Apex morphing (\ref{eq:A2A}) is used, between the given apex expressions of the categorical dataset:
\begin{equation}
\label{eq:A2A}
\text{Apex to Apex}
\begin{cases}
\mathbf{F}_{i}^{A_{1}rA_{2}}=M_{p}\left(\mathbf{F}_{i}^{A_{1}}, \mathbf{F}_{i}^{A_{2}}, r\right) \\
I_{i}^{A_{1}rA_{2}}=(1-r) I_{i}^{A_{1}} + r I_{i}^{A_{2}}\\
\theta^{A_{1}rA_{2}}=(1-r) \theta^{A_{1}} + r \theta^{A_{2}}
\end{cases}
\end{equation}
\noindent where $A$, $A_{1}$ and $A_{2}$ are apex expressions from the parent dataset.
In order to synthesize new intensity variations, Neutral to Apex morphing (\ref{eq:N2A}) is used, between the NE image and a given (or interpolated) apex image:
\begin{equation}
\label{eq:N2A}
\text{Neutral to Apex}
\begin{cases}
\mathbf{F}_{i}^{rA}=M_{p}\left(\mathbf{F}_{i}^{\textrm{Ne}}, \mathbf{F}_{i}^{A}, r\right) \\
I_{i}^{rA}=r I_{i}^{A} \\
\theta^{rA}=\theta^{A}
\end{cases}
\end{equation}

Fig.~\ref{fig:morphing_examples} shows an example of these 2 types of morphing. Once new interpolated apex expressions are generated by equation (\ref{eq:A2A}), `neutral to interpolated apex' morphings can further be generated by applying equation (\ref{eq:N2A}) on them. 

For every given or generated face image $\mathbf{F}_{i}^{E}$, with $I_{i}^{E}$ and $\theta^{E}$, the Valence and Arousal annotations can be computed as $V_{i}^{E}=I_{i}^{E}\cos(\theta^{E})$ and $A_{i}^{E}=I_{i}^{E}\sin(\theta^{E})$.

\begin{figure}[htb]
  \centering
  \includegraphics[width=\columnwidth]{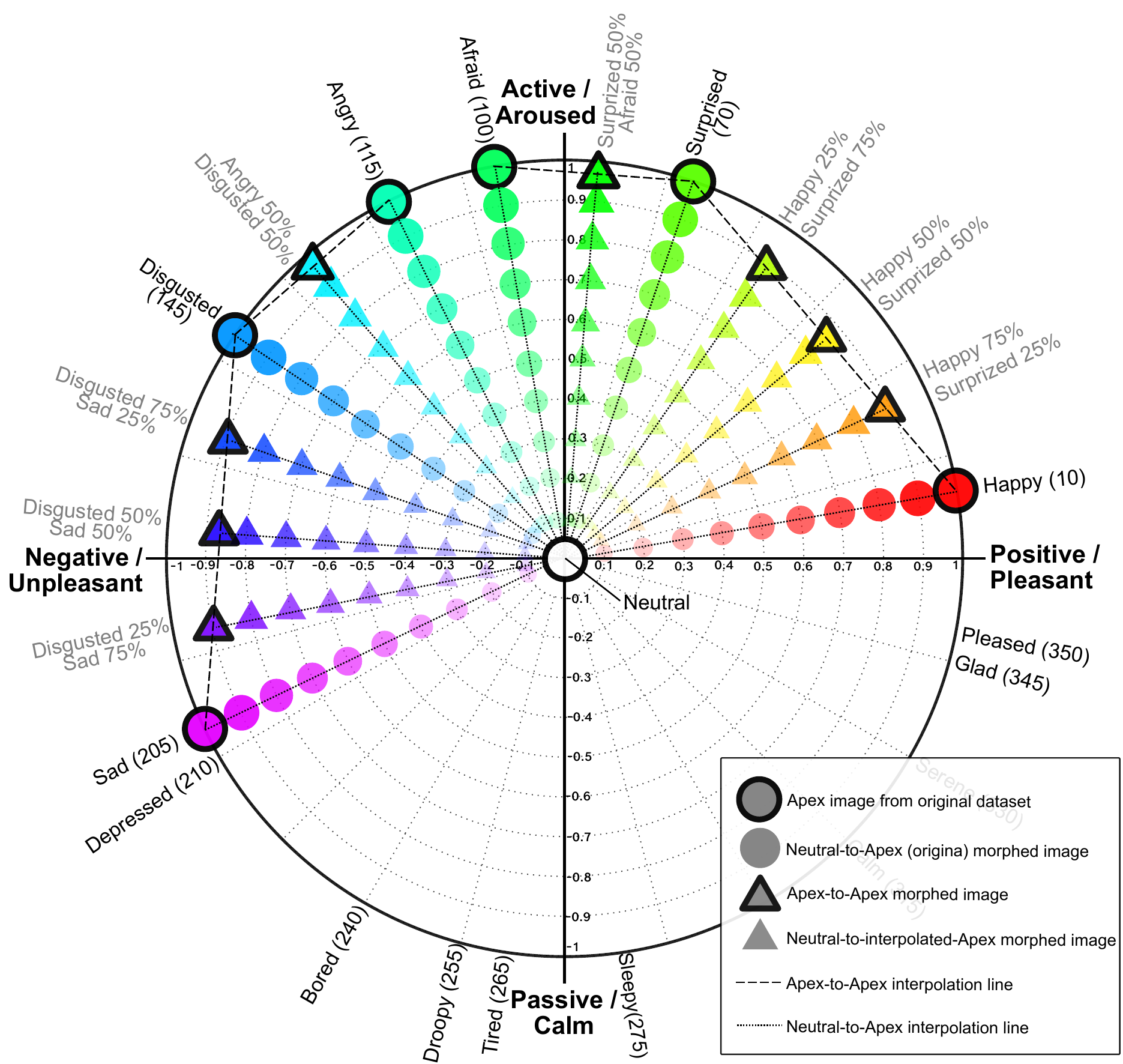}
	\caption{Dataset augmentation framework based on face morphing. Intensity of expression is represented with size and color saturation. 
	Outlined shapes indicate apex expressions.}
    \label{fig:AVspace}
\end{figure}

Fig.~\ref{fig:AVspace} illustrates the proposed augmentation framework for a typical categorical dataset with the 7 prototypical expressions. We start from $10^{\circ}$, the approximate location of Happy in VA space, and proceed in increments of $15^{\circ}$ steps in order to span the whole range up to $205^{\circ}$, where Sad is approximately located. The proposed template is bounded within $[10^{\circ},205^{\circ}]$ only because negative Arousal expressions (Sleepy, Tired, Bored, Calm, etc.) are absent from the typical categorical emotion datasets. We use an angle increment of $15^{\circ}$ and an intensity increment of $0.1$, because they strike a good balance between expression granularity, augmentation factor and symmetry between the positions of the given prototypical expressions in the AV space. 

Based on the above selected granularity, for a typical categorical dataset of 7 facial images per subject, the proposed augmentation framework can generate 134 new images, reaching a total of 134+7=141 facial images per subject. This translates to an augmentation factor of 20x, or 40x with simple image mirroring. Doubling the angular and radial granularity to $7.5^{\circ}$ and $0.05$, respectively, results in an augmentation factor of 80x, or 160x with image mirroring.

We build an example augmented dimensional dataset from a combination of 3 categorical datasets, which have been extensively used in psychology: the Radboud \cite{rafd},  Karolinska \cite{karolinska}, and  Warsaw \cite{warsaw} datasets. We select them because they have superior image quality, their facial expressions were guided by FACS experts, and they are accompanied by validation studies that provide the \emph{perceived} intensity of expression $I_{i}^{E}$. 


Using the proposed MorphSet augmentation framework, all faces are aligned with respect to the subjects' eyes, and synthetic images are generated for each subject. The resulting augmented dataset comprises more than 55,000 images, which -- with finer granularity and mirroring -- can reach over 450,000. More importantly though, each image comes with continuous Valence, Arousal, and Intensity annotations, which can be used to train dimensional FEA systems.

As the proposed augmented dataset contains only frontal images, an FEA system trained on it would not be invariant to different head poses. There are various solutions to this, which are discussed in detail in \cite{ICIP2020}.  In fact, we have successfully used such an approach to build a robust in-the-wild facial expression analysis system \cite{arXiv2021}.


\begin{figure*}[htb]
  \centering
  \includegraphics[width=\textwidth]{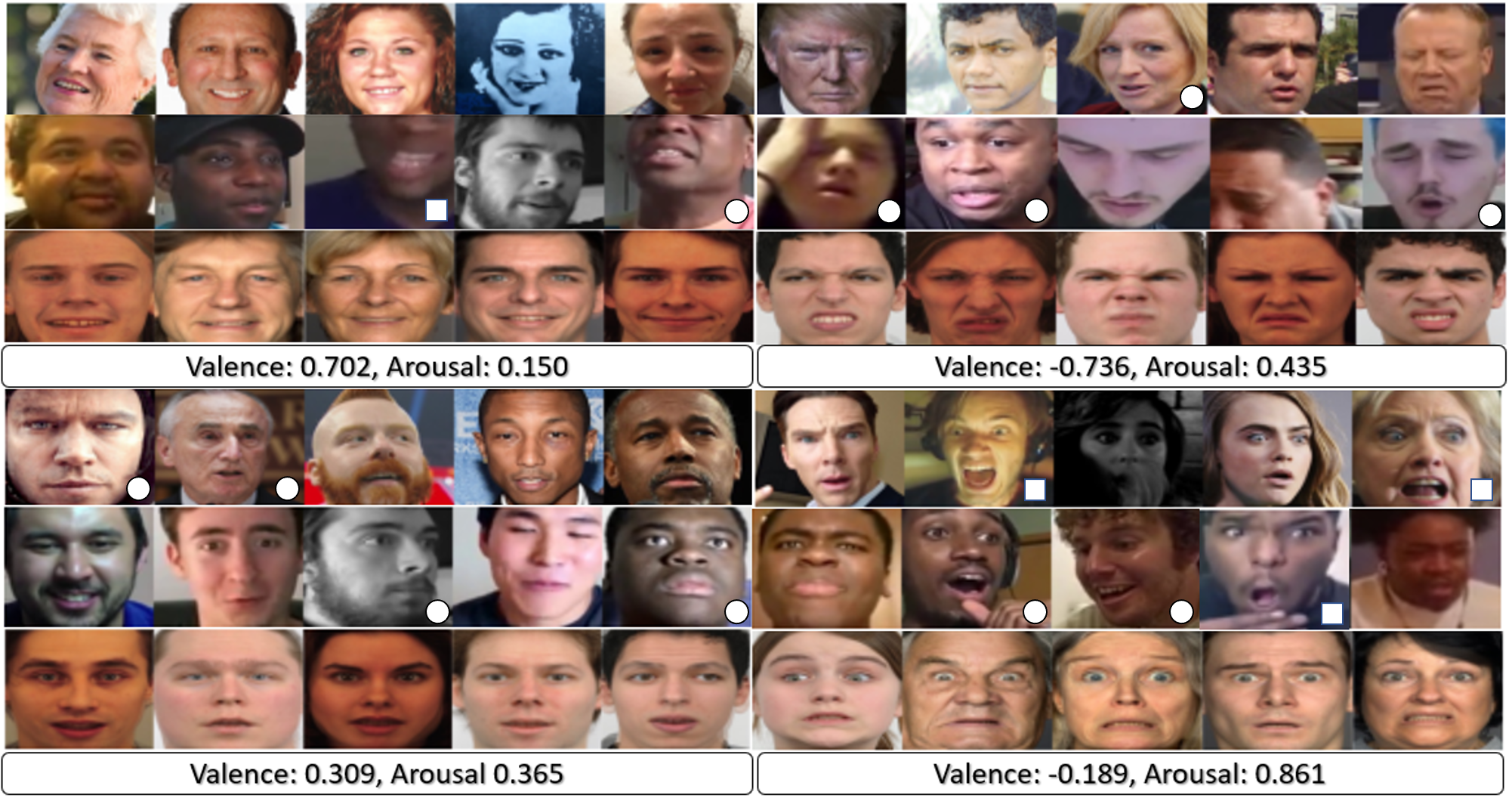}
	\caption{Randomly sampled training images around the same valence and arousal annotations from AffectNet, Aff-Wild, and MorphSet (top to bottom rows). Circles indicate wrong Arousal/Valence annotations (e.g.\ positive instead of negative), while squares indicate correct overall emotion but wrong intensity (e.g.\ extremely afraid instead of slightly afraid).}
     \label{fig:sample_4_classes}
\end{figure*}

\section{Comparison \& Discussion}
\label{sec:results}

In this section, we evaluate the benefits of MorphSet as a dataset for facial emotion recognition. We also describe the AffectNet \cite{affectnet} and Aff-Wild \cite{Aff-Wild} databases commonly used for emotion recognition. 


AffectNet \cite{affectnet} is a dataset comprising of roughly 450,000 images of in-the-wild-facial expressions collected from the Internet using affect keywords. These images are manually annotated with both categorical and dimensional labels. Despite its size, the training samples of AffectNet are noisy and highly imbalanced, causing many learning algorithms to perform poorly on the minority classes. Furthermore, human annotator agreement is just over 60\%, suggesting that the dataset suffers from noisy/incorrect annotations. 


Aff-Wild \cite{Aff-Wild} is an in-the-wild video dataset, consisting of 298 Youtube videos displaying reactions of 200 subjects. 
The annotations for valence and arousal were collected continuously via joystick. 
For direct comparison with AffectNet and MorphSet, we extract individual frames according to the affect annotations from the time-stamped video sequences. 

Table~\ref{tab:stats} compares these two databases with MorphSet, which is unique in that it provides a balanced distribution of facial expressions for each subject.

\begin{table}[htb]
\centering
\caption{Comparison to other dimensional datasets.}
\small
\begin{tabular}{cccc}
\hline
\textbf{}  
& AffectNet & Aff-Wild & \textbf{MorphSet}\\ \hline
\begin{tabular}[c]{@{}c@{}}Unique subjects\end{tabular}            & $\approx$450,000  
&200&167            \\ 
\begin{tabular}[c]{@{}c@{}}Expressions \\ per subject\end{tabular} & 1  & \begin{tabular}[c]{@{}c@{}}N/A \\ (video)\end{tabular}& $\approx$\textbf{342}         \\ 
\begin{tabular}[c]{@{}c@{}}Total \\ images\end{tabular}  & $\approx$450,000 &1,224,100       & \begin{tabular}[c]{@{}c@{}}55,000 up to\\  $\approx$450,000\end{tabular}  \\  
\begin{tabular}[c]{@{}c@{}}Annotators \end{tabular}  &12 &8 &  N/A        \\  \hline
\end{tabular}
\label{tab:stats}
\end{table}



To compare baseline FEA results, we train a  ResNet-18 model \cite{he2015deep} on each dataset.  We add two neurons after the final fully connected layer to predict valence and arousal dimensions and minimize the L2 loss.
Input images are resized to 224x224 with standard augmentations (e.g.\ affine transformations). 
We evaluate on the validation set of AffectNet and on 20\% of randomly selected, unseen identities of Aff-Wild and MorphSet. 

\balance

Table~\ref{tab:results} shows the Root Mean Square error (RMSE) and Concordance Correlation Coefficient (CCC) for each dataset. Although the results are not directly comparable because of the difference in test sets, they provide useful insights. The performance for MorphSet is significantly better than AffectNet and Aff-Wild, likely due to the frontal and highly controlled conditions of the images as well as higher consistency of the VA annotations. 

The fact that wild datasets are often noisier and less controlled in terms of facial expressions than MorphSet is further illustrated by Fig.~\ref{fig:sample_4_classes}, where images with specific VA labels are randomly sampled from each of the three datasets.  
AffectNet and Aff-Wild samples show significant fluctuations and outliers in facial expressions (indicated with circles and squares in Fig. \ref{fig:sample_4_classes}), whereas the facial expressions from MorphSet are much more consistent across different subjects.  

\begin{table}[htb]
\centering
\caption{ResNet-18 baseline results for each dataset.}
\begin{tabular}{rccc}
\hline
 & AffectNet & Aff-Wild  & \textbf{MorphSet} \\ \hline
\begin{tabular}[c]{@{}c}RMSE Valence \end{tabular} &0.427 &0.407 &0.157 \\ 
\begin{tabular}[c]{@{}c}RMSE Arousal \end{tabular} &0.390 &0.266 &0.155\\ \hline
\begin{tabular}[c]{@{}c}CCC Valence \end{tabular}  &0.533 &0.186 &0.915 \\  
\begin{tabular}[c]{@{}c}CCC Arousal \end{tabular}  &0.418 &0.174 &0.821 \\ \hline

\end{tabular}
\label{tab:results}
\end{table}

\section{Conclusions}
\label{sec:page}

We presented MorphSet, a dataset augmentation framework that can transform a typical categorical dataset of facial expressions into a balanced augmented one using image morphing. We use the framework to generate hundreds of different expressions per subject with full user control over their distribution. The augmented dataset comes with automatically generated, highly consistent dimensional annotations suitable for supervised learning of continuous affect.



\newpage
\balance
\small
\bibliographystyle{IEEEtran}
\bibliography{short,refs}

\end{document}